\documentclass{article}

    \usepackage[preprint]{neurips_2021}

\usepackage[utf8]{inputenc} 
\usepackage[T1]{fontenc}    
\usepackage{hyperref}       
\usepackage{url}            
\usepackage{booktabs}       
\usepackage{amsfonts}       
\usepackage{nicefrac}       
\usepackage{microtype}      
\usepackage[dvipsnames]{xcolor}         
\usepackage{amsmath}
\usepackage{graphicx}
\usepackage{amsthm}

\setcitestyle{numbers,open={[},close={]},citesep={,}}
\DeclareMathOperator*{\argmin}{arg\,min}

\theoremstyle{definition}
\newtheorem{definition}{Definition}[section]

\def\SIMLR{{\sc si}ML{\sc r}}
\def\nI#1{\Delta I_{#1}}  
\def\eg{{\em e.g.},\ }
\def\ie{{\em i.e.},\ }

\long\def\comment#1{}

\title{\SIMLR: Machine Learning inside the SIR model for COVID-19 Forecasting}

\author{%
  Roberto Vega\\
  University of Alberta\\
  \texttt{rvega@ualberta.ca} \\
  \And
  Leonardo Flores \\
  Independent researcher\\
  \texttt{leonardo.flores.q@gmail.com} \\
  
  \And
  Russell Greiner \\
  University of Alberta\\
  \texttt{rgreiner@ualberta.ca} \\
}

\begin{document}

\maketitle

\begin{abstract}
  Accurate forecasts of the number of newly infected people during an epidemic are critical for making effective timely decisions.
  This paper addresses this challenge using the \SIMLR\ model, which incorporates machine learning (ML) into the epidemiological SIR model. 
For each region,
\hbox{\SIMLR}\ 
  {} tracks the changes in the policies implemented at the government level, 
  which it uses to estimate the time-varying parameters of an SIR model for forecasting the number of new infections 
  1- to 4-weeks in advance.
  It also forecasts the probability of changes in those government policies at each of these future times, 
  which is essential for the longer-range forecasts. We applied \SIMLR\ to data from regions in Canada and in the United States,
  and show that its MAPE (mean average percentage error) performance is
  as good as SOTA forecasting models, with the added advantage of being an interpretable model.
  We expect that this approach will be useful not only for forecasting COVID-19 infections,
  but also in predicting the evolution of other infectious diseases.
\end{abstract}

\section{Introduction}\label{sec:introduction}

Since its identification in December 2019, COVID-19 has posed critical challenges for the public health and economies of 
essentially every country in the world~\cite{dong2020interactive,fauci2020covid,liu2020covid}. 
Government officials have taken a wide range of measures in an effort to contain this pandemic, 
including closing schools and workplaces, setting restrictions on air travel, and establishing stay at home requirements~\cite{hale2021global}. 
Accurately forecasting the number of new infected people in the short and medium term is critical for the timely 
decisions about policies and for the proper allocation of medical resources~\cite{arik2020interpretable,liao2020tw}.
For the general population (\eg business owners, parents with children in school age), 
additionally forecasting when the government will take action might help them to plan accordingly.

There are three basic approaches for predicting the dynamics of an epidemic:
compartmental models, statistical methods, and ML-based methods~\cite{arik2020interpretable,watson2021pandemic}. 
{\bf Compartmental models}\ subdivide a population into mutually exclusive categories (compartments), with a set of dynamical equations that explain the transitions among categories~\cite{blackwood2018introduction}. 
The \emph{Susceptible-Infected-Removed} (SIR) model~\cite{kermack1927contribution} is a common choice for the modelling of infectious diseases.
{\bf Statistical methods}, as the name implies, extract general statistics from the data to fit mathematical models that explain the evolution of the epidemic~\cite{liao2020tw}. 
Finally, {\bf ML-based methods}\ use machine learning algorithms to analyze historical data and find patterns that lead to accurate predictions of the number of new infected people~\cite{murphy2012machine,watson2021pandemic}.

Arguably, when any approach is used to make high-stake decisions,
it is important that it be not just accurate, 
but also interpretable: 
It should give the decision-maker enough information to justify the recommendation~\cite{rudin2019stop}. 
Here, we propose \SIMLR, which is an interpretable probabilistic graphical model (PGM) that combines compartmental models and ML-based methods. As its name suggests, it incorporates machine learning (ML) within an SIR model.

\SIMLR\ uses a \emph{mixture of experts}\ approach~\cite{jacobs1991adaptive},
where the contribution of each \emph{expert} to the final forecast depends on 
the changes in the government policies implemented at various earlier time points. 
When there is no recent change in policies (2 to 4 weeks before the week to be predicted),
\SIMLR\ relies on an SIR model with time-varying parameters that are fitted using machine learning methods. 
When a change in policy occurs, 
\SIMLR\ instead relies on a simpler model that predicts that the new number of infections will remain constant. Note that forecasting the number of new infections 1 and 2 weeks in advance
($\nI{1}$\ and $\nI{2}$) is relatively easy as \SIMLR\ knows, at the time of the prediction,
whether the policy has changed recently.
However, for 3- or 4-week forecasts
($\nI{3}$\ and $\nI{4}$),
our model needs to estimate the likelihood of a future change of policy.
\SIMLR\ incorporates prior domain knowledge to estimate such policy-change probabilities.

This work makes three important contributions.
(1)~It empirically shows that an SIR model with time-varying parameters can describe the complex dynamics of COVID-19. 
(2)~It describes an interpretable model that predicts the new number of infections 1 to 4 weeks in advance,
achieving state-of-the-art results,
in terms of mean average percentage error (MAPE),
on data from regions from Canada and the United States.
(3)~It effectively 
estimates the probability of observing a change in the government policies, 1 to 4 weeks in advance. 

The rest of Section~\ref{sec:introduction} describes the related work and basics of the SIR compartmental model. 
Section~\ref{sec:SIMLR} then describes in detail our proposed \SIMLR\ approach.
Finally, Section~\ref{sec:Experiments} shows the results of the predicting the number of new infections in the United States and provinces of Canada. 

\subsection{Basic SIR model}\label{sec:fixed_SIR} 
The \emph{Susceptible-Infected-Removed} (SIR) compartmental model~\cite{kermack1927contribution} is a mathematical model of infectious disease dynamics that divide the population into 3 disjoint groups~\cite{blackwood2018introduction}: \emph{Susceptible (S)} refers to the set of people who have never been infected, but can acquire the disease if they interact with someone with the infection. 
\emph{Infected (I)} refers to the set of people who have and can transmit the infection at a given point in time. 
\emph{Removed (R)} refers to the people who have either recovered or died from the infection,
and so cannot transmit the disease anymore. The transitions between the groups is given by the following differential equations:
\begin{equation}\label{eq:SIR}
    \frac{dS}{dt}\ =\ - \beta\ \frac{ S(t) I(t) }{N}
     \qquad\quad
        \frac{dI}{dt}\ =\ \beta \ \frac{S(t) I(t) }{N} - \gamma I(t)
        \qquad\quad
        \frac{dR}{dt}\ =\ \gamma\ I(t)
\end{equation}

\begin{figure}
  \centering
  \includegraphics[width=\columnwidth]{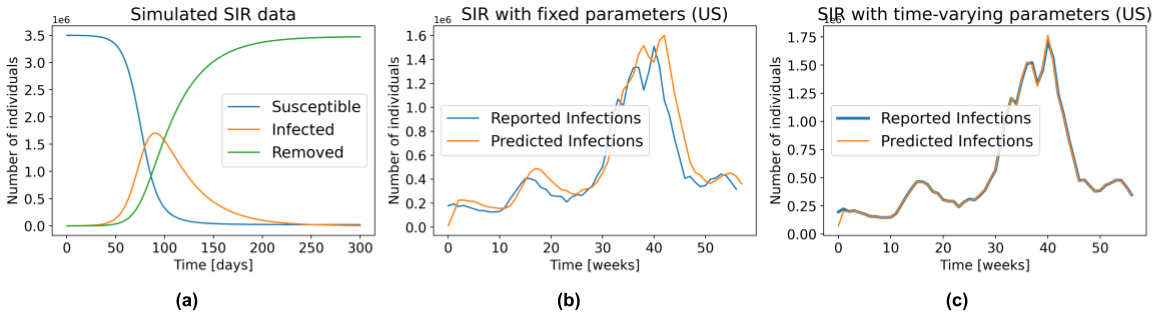}
  \caption{\label{fig:SIR}
  \textbf{(a)} General behaviour of the SIR model.
  \textbf{(b)} The number of infections predicted by the SIR model with fixed parameters,
  fitted to the US data for 1 week prediction.
  \textbf{(c)} Similar to (b), but with time-varying parameters.
  }
\end{figure}

This simple model, which assumes homogeneous and constant population, is fully defined by the parameters $\beta$ (transmission rate) and $\gamma$ (recovery rate). The intuition behind this model is that every infected patient gets in contact with $\beta$ people. 
Since only the susceptible people can become infected, the chance of interacting with a susceptible person is simply the proportion of susceptible people in the entire population, $N=S+I+R$. 
Likewise, at every time point,
$\gamma$ proportion of the infected people is removed from the system. 
Figure~\ref{fig:SIR}(a) depicts the general behaviour of an SIR model.

\subsection{Related work} 
\label{sec:RelatedW}

The main idea behind combining compartmental models with machine learning is to replace the fixed parameters of the former with time-varying parameters that can be learned from data~\cite{anastassopoulou2020data,arik2020interpretable,chen2020time,liao2020tw,liu2012infectious,watson2021pandemic}. 
However, it is still not clear if 
these approaches are able to explain (and forecast) the different waves of COVID-19 infections.

Particularly relevant to our approach is the work by \citet{arik2020interpretable}, who used latent variables and autoencoders to model extra compartments in an extended \emph{Susceptible-Exposed-Infected-Removed}~(SEIR) model~\cite{blackwood2018introduction}. 
Those additional compartments bring further insight into how the disease impacts the population; however, we argue that they are not needed for an accurate prediction of the number of new infections; see results in Section~\ref{sec:Experiments}. 
One limitation of their model is a decrease in performance when the trend in the number of new infections changes~\cite{arik2020interpretable}.
We hypothesize that those changes in trend are related to the government policies that are in place at a specific point in time. 
\SIMLR\ is able to capture those changes by tracking the policies implemented at the government level, 
allowing it to also predict when 
those policies are likely to change.

A different line of work uses machine learning methods to directly predict the number of new infections,
without incorporating epidemiological  models~\cite{jin2021inter,kafieh2021covid,mojjada2020machine,omran2021applying}. Like our approach,
\citeauthor{yeung2021machine} added the non-pharmaceutical interventions (policies) as features in their models; 
then they used traditional machine learning models to make their predictions under the assumption that the data is not a time-series, 
but a collection of points that are independent and identically distributed~\cite{yeung2021machine}. 
Their approach is limited to make predictions up to 2 weeks in advance, since information about the policies that will be implemented in the future is not available at inference time.
Our \hbox{\SIMLR}{} approach differs by being interpretable and also forecasting policy changes, which allows it to extend the horizon of the $\nI{}$ predictions.

\section{Our \SIMLR\ model}\label{sec:SIMLR}
\begin{figure}[tb] 
  \centering
  \includegraphics[width=\columnwidth]{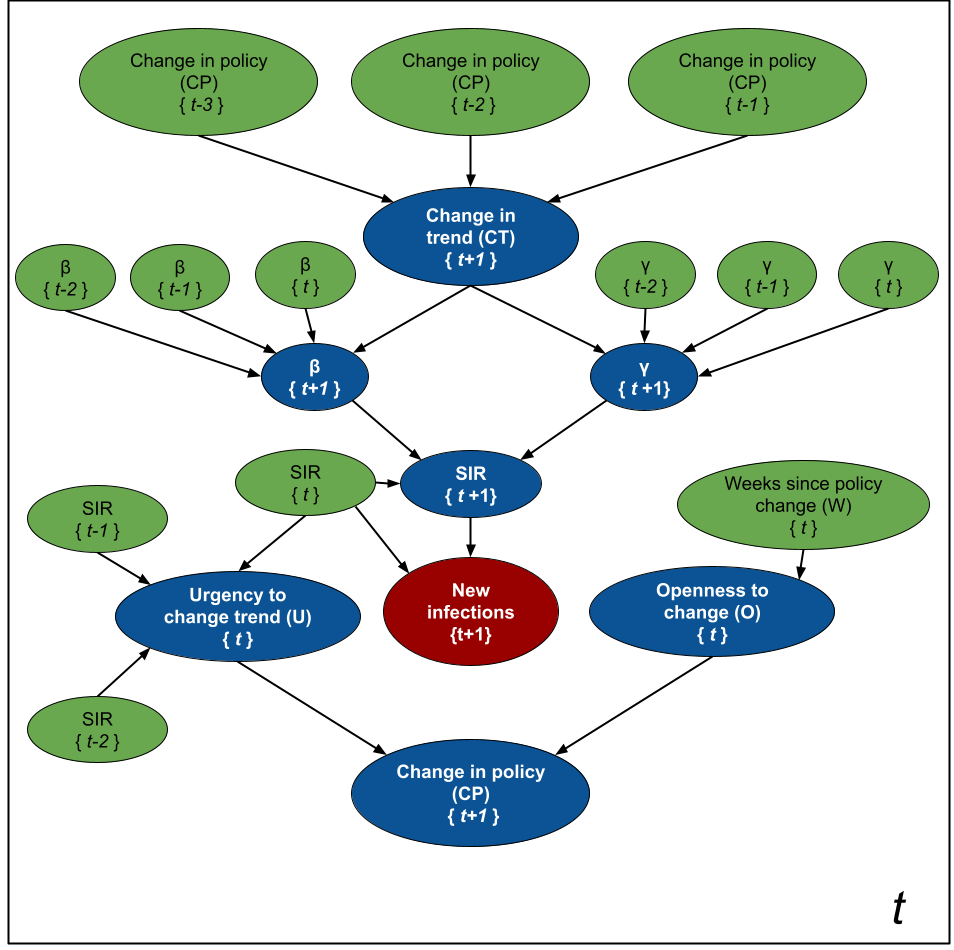}
  \caption{\label{fig:SIMLR} Modeling \SIMLR\ as a probabilistic graphical model for forecasting new cases of COVID-19. 
  The {\color{blue}\bf blue} nodes are estimated at each time point, while the 
  {\color{OliveGreen}\bf green} ones are either based on past information, or where estimated in a previous iteration.}
\end{figure}

We view \SIMLR\ as a probabilistic graphical model that uses a mixture of experts approach to forecast the number of new COVID-19 infections, 1 to 4 weeks in advance; Figure~\ref{fig:SIMLR} depicts this graphical model 
as 
a plate model~\cite{koller2009probabilistic}. 
The {\color{blue}\bf blue} nodes 
are {\em estimated}\ at every time point,
while the values of the 
{\color{OliveGreen}\bf green} nodes 
are either {\em known}\ 
as part of the historical data, 
or 
{\em inferred}\ in a previous time point. 

The random variables in Figure~\ref{fig:SIMLR} are assumed to have the following distributions:
\begin{equation}
    \begin{array}{r@{\quad |\quad }c@{\quad \sim\quad }l}
    \mbox{CT}_{t+1} & 
      \mbox{CP}_{t-3},\ \mbox{CP}_{t-2},\  \mbox{CP}_{t-1}& 
      Categorical_{K \in \{-1,0,1\}} (\theta_{CT})\\
    \beta_{t+1} & \beta_{t-2},\ \beta_{t-1},\ \beta_{t},\ \mbox{CT}_{t+1} & \mathcal{N}(\mu_{\beta}, \Sigma_\beta)\\
    \gamma_{t+1} &\gamma_{t-2},\ \gamma_{t-1},\ \gamma_{t},\ \mbox{CT}_{t+1} & \mathcal{N}(\mu_{\gamma}, \Sigma_\gamma)\\
    \mbox{SIR}_{t+1} &\beta_{t+1},\ \gamma_{t+1} & \mathcal{N}(\mu_{SIR}, \Sigma_{SIR})\\
    U_{t} &\mbox{SIR}_{t-2}, \mbox{SIR}_{t-1},\ \mbox{SIR}_{t}& Categorical_{K \in \{-1,0,1\}} (\theta_{U})\\
    O_{t}&W_t & Categorical_{K \in \{0,1\}} (\theta_{O})\\
    \mbox{CP}_{t+1}&O_{t},\ U_t & Categorical_{K \in \{-1,0,1\}} (\theta_{CP})
    \end{array}
\end{equation}

\noindent where $t$ indexes the current week,
$SIR_t = [S_t, I_t, R_t]$,
$\mu_{SIR} \in \mathbb{R}^3$ is given below by Equation~\ref{eq:SIR_discrete}$, \mu_\beta = (\alpha_{0,CT_{t+1}}) + (\alpha_{1,CT_{t+1}})\beta_{t-1} + (\alpha_{2,CT_{t+1}})\beta_{t-2} + (\alpha_{3,CT_{t+1}})\beta_{t-3}$ and  $\mu_\gamma = (\omega_{0,CT_{t+1}}) + (\omega_{1,CT_{t+1}})\gamma_{t-1} + (\omega_{2,CT_{t+1}})\gamma_{t-2} + (\omega_{3,CT_{t+1}})\gamma_{t-3}$ are linear combinations of the three previous values of $\beta$ and $\gamma$ (respectively). 
The coefficients of those linear combinations depend on the value of the random variable $\mbox{CT}_{t+1}$. We did not specify a distribution for the node $\mbox{\tt New infections}_{t+1}$ because its value is deterministically computed as $S_t - S_{t+1}$.

Informally, the assignment $\mbox{CT}_t = -1$ means that we expect a change in trend from an increasing number of infections to a decreasing one. The opposite happens when $\mbox{CT}_t = 1$, 
while $\mbox{CT}_t = 0$ means that we expect the population to follow the current trend (either increasing or decreasing). 
These changes in trend are assumed to depend on changes in the government policies 2 to 4 weeks prior to the week of our forecast -- \hbox{\eg}{} we use $\{\mbox{CT}_{t-3},\, \mbox{CT}_{t-2}, \,\mbox{CT}_{t-1}\}$
when predicting the number of new infections at $t+1$, $\nI{t+1}$;
and we need
$\{\mbox{CT}_{t},\, \mbox{CT}_{t+1}, \,\mbox{CT}_{t+2}\}$
when predicting $\nI{t+4}$,
Note that, at time $t$, we will not know $\mbox{CT}_{t+1}$ nor $\mbox{CT}_{t+2}$.

The status of $\mbox{CT}_{t+1}$ defines the coefficients that relate $\beta_{t+1}$ and $\gamma_{t+1}$ with their three previous values $\beta_{t}, \beta_{t-1}, \beta_{t-2}$ and $\gamma_{t}, \gamma_{t-1}, \gamma_{t-2}$, respectively. 
Since $\beta_{t+1}$ and $\gamma_{t+1}$ fully parameterize the SIR model in Equation~\ref{eq:SIR},
we can estimate the new number of infected people, $\nI{t+1}$, from these parameters (as well as the SIR values at time $t$).

The random variables $U_t \in \{-1, 0, 1\}$ and $O_t \in \{0, 1\}$ are auxiliary variables
designed to predict the probability of observing a change in policy at time $t+1$. Intuitively, $U_t$ represents the ``urgency'' of modifying a policy. 
As the number of cases per 100K inhabitants and
the rate of change between the number of cases in two consecutive time points increases, 
the urgency to set {\em stricter}\ government policies increases.
As the number (and rate of change) of cases decreases, 
the urgency to {\em relax}\ the policies increases. 
Finally, $O_t$ models the ``willingness'' to execute a change in government policies. 
As the number of time points without a change increases,
so does this ``willingness''.

\subsection{SIR with time-varying parameters}\label{sec:SIR_time_varying} 
We can approximate an SIR model by transforming the differential Equation~\ref{eq:SIR} into an equation of differences and then solving for $S_t, I_t$ and $R_t$:

\begin{equation}\label{eq:SIR_discrete}
   S_t\ =\ - \beta \frac{S_{t-1} I_{t-1} }{N} + S_{t-1}
   \qquad
        I_t \ =\ \beta\frac{ S_{t-1} I_{t-1} }{N} - \gamma I_{t-1} + I_{t-1}
    \qquad
        R_t\ =\ \gamma I_{t-1} + R_{t-1} 
\end{equation}

While the SIR model is non-linear with respect to the states (S, I, R), it is linear with respect to the parameters $\beta$ and $\gamma$. 
Therefore, under the assumption of constant and known population size  (\ie $N = S_t + I_t + R_t$)
we can re-write the set of Equations~\ref{eq:SIR_discrete} as:
\begin{equation}\label{eq:SIR_discrete_linear_algebra}
    \begin{aligned}
        \begin{bmatrix}
        S_t \\ I_t
        \end{bmatrix}\quad
        &= \quad
        \begin{bmatrix} -\frac{S_{t-1}I_{t-1}}{N} & 0 \\
        \frac{S_{t-1}I_{t-1}}{N}& - I_{t-1}\\
        \end{bmatrix}
        \begin{bmatrix}
        \beta \\ \gamma
        \end{bmatrix}\quad +\quad \begin{bmatrix}
        S_{t-1} \\ I_{t-1}
        \end{bmatrix}\\
        R_t \quad &= \quad N - S_t - I_t
    \end{aligned}
\end{equation}

Given a sequence of states $x_1, \dots, x_n$, where
$x_t\ =\ [S_t\ I_t]^T$, 
it is possible to estimate the optimal parameters of the SIR model as:
\begin{equation}\label{eq:beta_gamma}
    (\beta^*,\ \gamma^*)\quad =\quad \argmin_{\beta,\gamma} \sum_{i=1}^n||x_i - \hat{x}_i ||^2 + \lambda_1 (\beta - \beta_0)^2 + \lambda_2 (\gamma - \gamma_0)^2
\end{equation}
\noindent where $\hat{x}_i$ is computed using Equation~\ref{eq:SIR_discrete_linear_algebra},
and $\lambda_1$ and $\lambda_2$ are optional regularization parameters that allow the incorporation of the priors $\beta_0$ and $\gamma_0$. For the case of Gaussian priors
-- \ie $\beta \sim \mathcal{N}(\beta_0, \sigma_{\beta}^2)$ and $\gamma \sim \mathcal{N}(\gamma_0, \sigma_{\gamma}^2)$
-- we use $\lambda_1 = \frac{1}{2\sigma_{\beta}^2}$ and $\lambda_2 = \frac{1}{2\sigma_{\gamma}^2}$~\cite{murphy2012machine}.

In the traditional SIR model,
we set $\lambda_1 = \lambda_2 = 0$ and fit a single $\beta$ and $\gamma$  to the entire time series. 
However, as shown in Figure~\ref{fig:SIR}(a), 
an SIR model with fixed parameters is unable to accurately model 
several waves of infections. 
As illustration,
Figure~\ref{fig:SIR}(b) shows the predictions produced by fitting an SIR with fixed parameters (Equation~\ref{eq:beta_gamma}) to the US data from 
29/March/2020 to 3/May/2021,
and then using those parameters to make predictions one week in advance,
{over this same interval}.
That is, using this learned $(\beta, \gamma)$,
and the number of people in the $S$, $I$, and $R$ compartments on 28/March/2020,
we predicted the number of observed cases during the week of 29/March/2020 to 4/April/2020. 
Then, using 
the reported $[S, I, R]$ values on 4/April/2020,
we predicted the number of new infections in the week of
5/April/2020 to 11/April/2020 and so on. Note that even though the parameters $\beta$ and $\gamma$ were found using {\em the entire time series}
-- \ie using information that was not available at the time of prediction --
the resulting model still does a poor job fitting the reported data.

Figure~\ref{fig:SIR}(c), on the other hand, was created by allowing $\beta$ and $\gamma$ to change every week. 
Here, we first found the parameters that fit the data from 
29/March/2020 to 4/April/2020
-- call them $\beta_1$ and $\gamma_1$ --
then used those parameters along with the SIR state on 
28/March/2020 to predict the number of new infections one week ahead. By repeating this procedure during the entire time series we obtained an almost perfect fit to the data. 
Of course, these are 
also
not ``legal'' predictions since they 
too
use information that is not available at prediction time
-- \ie they used the number of reported infections during this first week to find the parameters, which were then used to estimate the number of cases over this time.
However, this ``cheating'' example shows that
 an SIR model, with the best time varying parameters, can model the complex dynamics of COVID-19, including the different waves of infections. 
Recall from Figure~\ref{fig:SIR}(b) that this is not the case in the SIR model with fixed parameters:
even in the best scenario 
(knowing the optimal parameters based on the true values over months of the same data you want to fit!),
it still cannot properly fit the data.

\subsection{Estimating $\beta_{t+1}$ and $\gamma_{t+1}$}
\label{sec:parameterEstimation} 

Naturally, the challenge is ``legally'' computing the appropriate values of
$\beta_{t+1}$ and $\gamma_{t+1}$, for each week,
using only the data that is known at time $t$.
Figure~\ref{fig:SIMLR} shows that computing $\beta_{t+1}$ and $\gamma_{t+1}$ depends on the status of the random variable $\mbox{CT}_{t+1}$. 
When $\mbox{CT}_{t+1} = 0$ 
 -- \ie there is no change in the current trend --
 we assume that:
\begin{equation}\label{eq:beta_gamma_distribution}
    \begin{gathered}
    \beta_{t+1}\quad \sim\quad \mathcal{N}(\alpha_0 + \alpha_1 \beta_{t} + \alpha_2 \beta_{t-1} + \alpha_3 \beta_{t-2},\quad \sigma_{\beta}^2)\\
    \gamma_{t+1}\quad  \sim \quad \mathcal{N}(\omega_0 + \omega_1 \gamma_{t} + \omega_2 \gamma_{t-1} + \omega_3 \gamma_{t-2}, \quad \sigma_{\gamma}^2)
    \end{gathered}
\end{equation}
 
 At time \emph{t},
 we can use the historical \emph{daily} data $x_1, x_2, \dots, x_{t}$ to find the \emph{weekly} parameters 
 $\beta_1, \beta_2, \dots, \beta_{t/7}$ and $\gamma_1, \gamma_2, \dots, \gamma_{t/7}$. The first {\em weekly}\ pair $(\beta_1, \gamma_1)$ is found  by fitting Equation~\ref{eq:beta_gamma} to $x_1, \dots, x_7$; 
$(\beta_2, \gamma_2)$ to $x_8, \dots, x_{14}$; and so on. 
Finally, we find the parameters $\boldsymbol{\alpha}$ and $\boldsymbol{\omega}$ in Equation~\ref{eq:beta_gamma_distribution} by maximizing the likelihood of the computed pairs. After finding those parameters, it is straightforward to infer $(\beta_{t+1}, \gamma_{t+1})$~\cite{koller2009probabilistic}.

Note that we decided to estimate $\beta_{t+1}$ and $\gamma_{t+1}$ as a function of the 3 previous values of those parameters;
this allows them to incorporate the velocity and acceleration at which the parameters change. 
That is, we can estimate the velocity at which $\beta$ change as $v_{\beta,t} = \beta_t - \beta_{t-1}$ and its acceleration as $a_{\beta,t} = v_{\beta,t} - v_{\beta,t-1}$.
Then, estimating $\beta_t = \theta_0 + \theta_1\beta_{t-1} + \theta_2 v_{\beta,t-1} + \theta_3 a_{\beta,t-1}$ is equivalent to the model in Equation~\ref{eq:beta_gamma_distribution}.
The same reasoning applies to the computation of $\gamma_t$.
We call this approach the ``trend-following varying-time parameters SIR'', tf-v-SIR.

For the case of $\mbox{CT}_t = -1$ and $\mbox{CT}_t = 1$ 
(which represents a change in trend from increasing number of infections to decreasing number of infections or vice-versa),
we set $\beta_{t+1}$ and $\gamma_{t+1}$ to values such that the predicted number of new cases at week $t+1$ is identical to the one at week $t$. We call this the ``Same as the Last Observed Week'' (SLOW) model.

\subsection{Estimating $\mbox{CT}_{t+1},\ \mbox{CP}_{t+1},\ \mbox{O}_t$
} 
\label{sec:Disc+Disc}
\begin{figure}[h]
  \centering
  \includegraphics[width=\columnwidth]{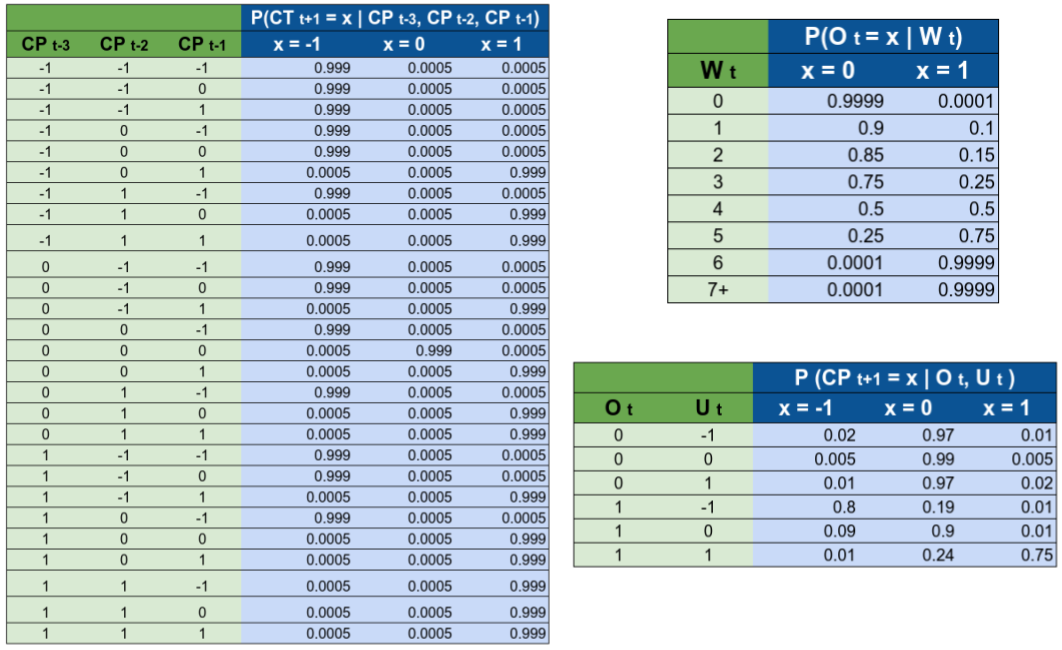}
  \caption{\label{fig:CPT}
  Conditional probability tables used by \SIMLR.
  The names of the variables refer to the nodes that appear on Figure~\ref{fig:SIMLR}}
\end{figure}

The random variables $\mbox{CT}_{t+1} \in\{-1,0,1\}, \mbox{CP}_{t+1}\in\{-1,0,1\}$ and $\mbox{O}_t\in\{0,1\}$ in Figure~\ref{fig:SIMLR} 
are all discrete nodes with discrete parents,
meaning  their probability mass functions are fully defined by conditional probability tables~(CPTs). 
Learning the parameters of such CPTs from data is 
challenging due to the scarcity of historical information. 
The random variable $\mbox{CT}_{t+1}$ depends on the random variable changes in policy ($\mbox{CP}$) at times $t-1, t-2, t-3$; 
however, there are very few changes in policy in a given region,
meaning it is difficult to accurately estimate those probabilities from data.
For the random variable $\mbox{O}$, which represents the ``willingness'' of the government to implement a change in policy, there is no observable data at all. 
We therefore relied on prior expert knowledge to set the parameters of the conditional probability tables for these random variables. 
Figure~\ref{fig:CPT} shows
the actual CPTs used in our experiments.

\subsection{Estimating $U_t$ 
} \label{sec:disc+cont}
For modelling the random variable $U_t$, which represents the ``Urgency to change the trend'',
we use an NN-CPD (neural-network conditional probability distribution), 
which is a modified version of the multinomial logistic conditional probability distribution~\cite{koller2009probabilistic}.

\begin{definition}[NN-CPD]
Let $Y \in \{1, \dots, m\}$ be an ${m}$-valued random variable 
with \emph{k} parents $X_1, \dots, X_k$ 
that each take on numerical values. 
The conditional probability distribution $P(Y\ |\ X_1,\dots,X_k)$ is an NN-CPD 
if there is an function $z\ =\ f_{\theta}(X_1, \dots, X_k)
\ 
\in \mathbb{R}^m$, 
represented as a neural network with parameters $\theta$, such that $p(Y = i \ |\ x_1, \dots, x_k) = \exp(z_i) / \sum_j \exp(z_j)$, where $z_i$ represents the \emph{i-th} entry of $z$.
\end{definition}

Similar to the case of discrete nodes with discrete parents, the main problem in estimating the parameters $\theta$ of the NN-CPD is the scarcity of data. 
To compensate for the lack of data,
we again rely on domain knowledge and use the probabilistic labels approach proposed by~\citet{vega2021sample}.
Using this approach, we can use a dataset with few training instances along with their probabilities to learn the parameters of a neural network more efficiently. 

To compute $P(U_{t} \ |\ \mbox{SIR}_{t-2}, \mbox{SIR}_{t-1},\ \mbox{SIR}_{t})$,
we extract two features: $c_t = \mbox{10E5} (S_{t-1} - S_{t}) / N$, which represents the  number of 
new infections 
per 100K inhabitants; and $v_t = c_t - c_{t-1}$, 
which estimates the rate of change of $c_t$,
then define 
$P(U_{t} \ |\ \mbox{SIR}_{t-2}, \mbox{SIR}_{t-1},\ \mbox{SIR}_{t}) = P(U_{t} \ |\ c_t, v_t)$.

\begin{figure}
  \centering
  \includegraphics[width=\columnwidth]{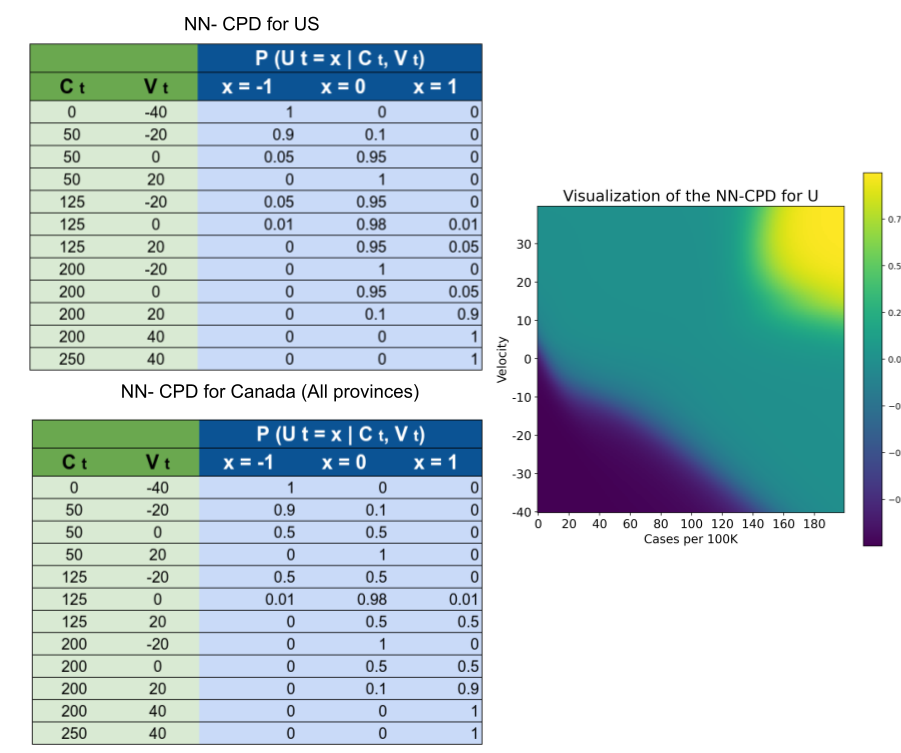}
  \caption{Dataset used to create the NN-CPD for the variable $U_t$ and its visualization. Values closer to 1 (yellow) increase $p(U_t = 1\ |\ C_t,V_t)$. Values closer to 0 (green) increase $p(U_t = 0\ |\ C_t,V_t)$. Values closer to $-1$ (blue) increase $p(U_t = -1\ |\ C_t,V_t)$ }\label{fig:NN_CPD}
\end{figure}

To learn the parameters $\theta$,
we created the dataset shown in Figure~\ref{fig:NN_CPD} and a neural network with a single hidden layer with 64 units with sigmoid activation functions, 
and 3 output units with softmax activation. 
As the number and rate of change of new infections increases, the ``urgency'' to impose stricter policies increases. The opposite effect (relaxing the policies) occurs with a decrease in the number and rate of change of new infections. When the new infections and its rate of change are relatively stable, then there is a preference to keep the current policies in place.

\section{Experiments and results}\label{sec:Experiments} 

We evaluated the performance of \SIMLR, in terms of MAPE, 
for forecasting the number of new infections 1 to 4 weeks in advance,
in data from United States (as a country) and the 6 biggest provinces of Canada: Alberta~(AB), British Columbia~(BC), Manitoba~(MN), Ontario~(ON), Quebec~(QB),
and Saskatchewan~(SK).
For each of the regions, the predictions are done on a weekly basis, over the 39 weeks from 26/Jul/2020 to 1/May/2021. This time span captures different waves of infections.

At the end of every week, we fitted the \SIMLR\ parameters using the data that was available
until that week.
For example, on 25/Jul/2020 we used all the data available from 1/Jan/2020 to 25/Jul/2020 to fit the parameters of \SIMLR. Then, we made the predictions for the new number of new infections during the weeks: 26/Jul/2020 -- 1/Aug/2020 (1 week in advance), 2/Aug/2020 -- 8/Aug/2020 (2 weeks in advance), 9/Aug/2020 -- 15/Aug/2020 (3 weeks in advance), and 16/Aug/2020 -- 22/Aug/2020 (4 weeks in advance). 
 After this, we then fitted the parameters with data up to 1/Aug/2020 and repeated the same process, for 38 more iterations, until we covered the entire range of predictions. 

We compared the performance of \SIMLR\ with the simple SIR compartmental model with time-varying parameters learned using Equation~\ref{eq:beta_gamma_distribution} but no other random variable (tf-v-SIR),
and with the simple model that forecasts that the number of cases 1 to 4 weeks in advance is the 
``Same as the Last Observed Week'' (SLOW).
For the United States data,
we also compared the performance of \SIMLR\ against the publicly available predictions at the COVID-19 Forecast Hub
({\tt https://github.com/reichlab/covid19-forecast-hub})~\cite{cramer2021evaluation}.

For training, we used the publicly available (Creative Commons Attribution CC BY standard) dataset OxCGRT~\cite{hale2021global},
which contains the policies implemented by different countries, as well as the time period over which they were implemented. 
We limited our analysis to 3 policy decision:
{\tt Workplace closing}, {\tt Stay at home requirements}, and {\tt Cancellation of public events}. 
For information about the new number of reported cases and deaths,
we used the publicly available (Creative Commons Attribution 4.0 International) COVID-19 Data Repository by the Center for Systems Science and Engineering at Johns Hopkins University ({\tt https://github.com/CSSEGISandData/COVID-19})~\cite{dong2020interactive}.

\subsection{Data preprocessing}
Before inputting the time-series data to \SIMLR, we performed some basic preprocessing: 
(1)~The original data contains the cumulative number of reported infections/deaths on a daily basis.
We trivially transformed this time-series into the number of new daily infections/deaths. 
(2)~We considered negative values from the new daily infections/deaths time-series as missing, 
assuming these negative values arose due to inconsistencies during the data reporting procedure. 
(3)~We ``filled-in'' the missing values. When the number of new infections/deaths was missing at day $d$, we assumed that the entry at $d+1$ contained the cases for both $d$ and $d+1$, and divided the number of new infections/deaths evenly between both days. 
(4)~We eliminated outliers. 
For each day $d$, with number of reported new infections, $\nI{d}$, we computed the mean $(\mu_d)$ and standard deviation $(\sigma_d$ of the set $\nI{d-10}, \dots, \nI{d-1}$;
we then set $\nI{d} := \min\{ \nI{d}, \mu_d + 4\sigma_d\}$.
(5)~We used the number of new infections and new deaths to produce the SIR vector $SIR_t = [S_t, I_t, R_t]$.
Here, we assume everyone in that region was susceptible at the start time -- \ie $S_{0} =$ total population.
At each new time, we transfer the number of new infections from $S$ to $I$, and the number of new deaths and recovered from $I$ to $R$.
If the number of new recovered people is not reported,
we assume that each new infected person that did not die within the next 15 days, has recovered.

\subsection{Results and discussion} 
\label{sec:Results}
\begin{figure} 
  \centering
  \includegraphics[width=0.9\columnwidth]{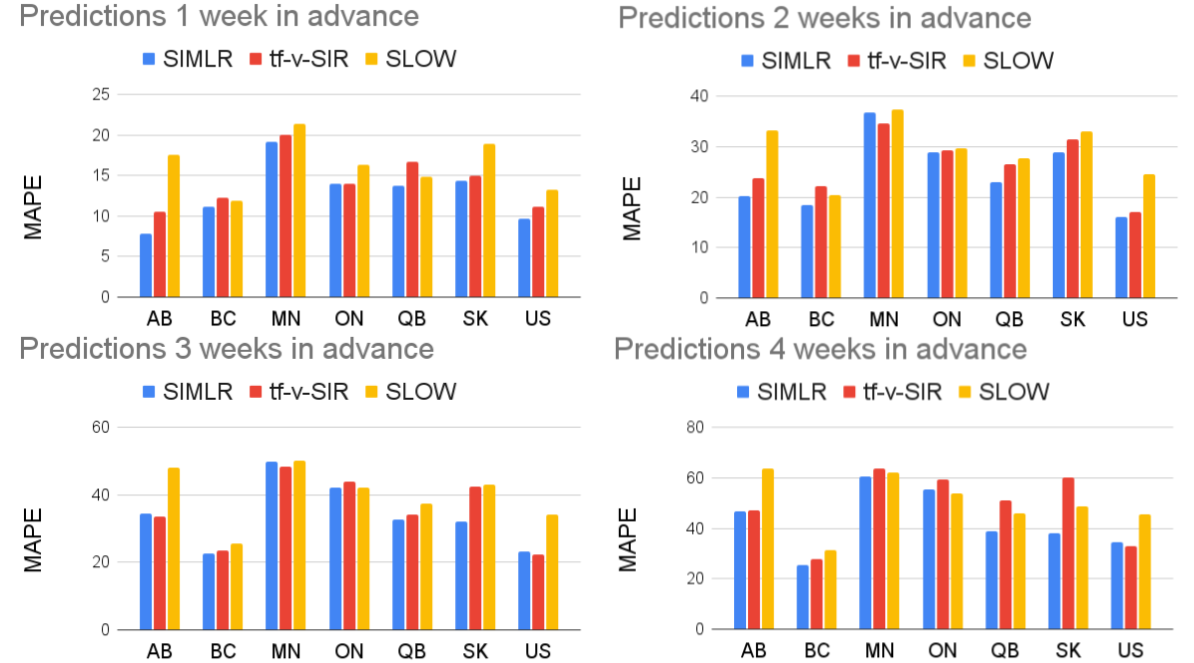}
  \caption{\label{fig:MAPE_Regions} 
  Comparison of \SIMLR, SIR model with time-varying parameters, and SLOW.}
\end{figure}

Figure~\ref{fig:MAPE_Regions} shows the MAPE of the 1- to 4-week forecasts for the United States (US) and 
the 6 biggest provinces of Canada. 
Note that \SIMLR\ has a consistently lower MAPE than tf-v-SIR and SLOW. 

\begin{figure}
  \centering
  \includegraphics[width=0.9\columnwidth]{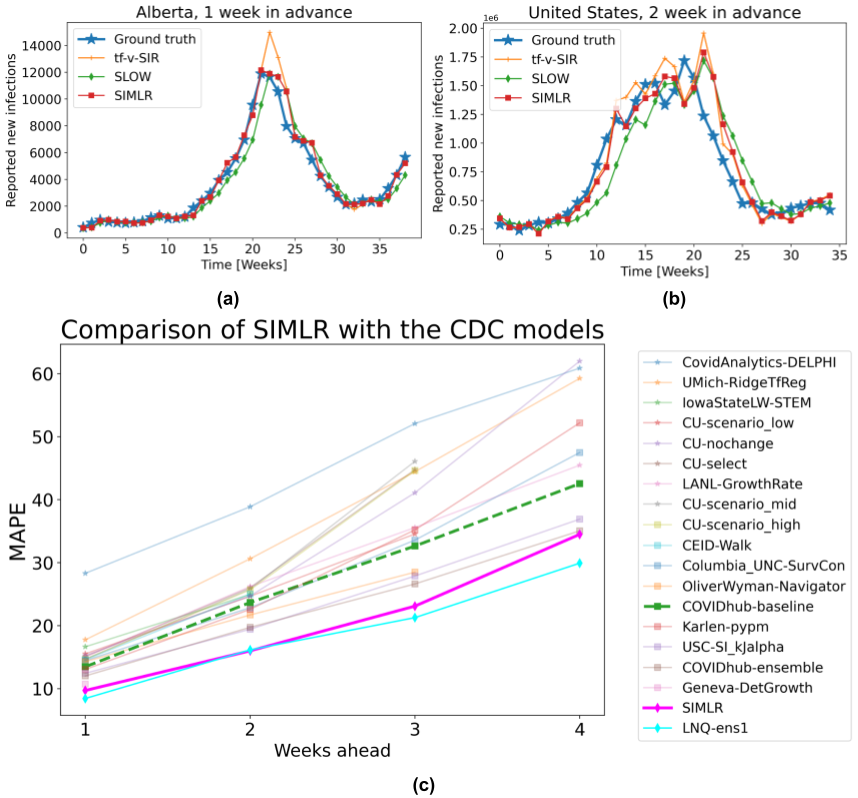}
  \caption{\label{fig:SIMLR_Results}
  (a)~1-week forecasts \SIMLR, tf-v-SIR, 
  and SLOW, for Alberta, Canada. 
  (b)~2-week forecasts, of the same models, for US data. (c)~Comparison of \SIMLR\ versus models submitted to the CDC (on US data). }
\end{figure}

Figure~\ref{fig:SIMLR_Results} illustrates the actual predictions of \SIMLR\ one week in advance (and tf-v-SIR and SLOW) for the province of Alberta, Canada; and two weeks in advance for the United States as a country. 
These two cases exemplify the behaviour of \SIMLR. 

As noted above, there is a 2- to 4-week lag after a policy changes,
before we see the effects.
This means the task of making 1-week forecasts is relatively simple,
as the relevant policy (at times $t-3$ to $t-1$) is fully observable. This allows \SIMLR\ to directly compute $\mbox{CT}_{t+1}$,
which can then choose whether to continue using the SIR with time-varying parameters if no policy changed at time $t-1$, $t-2$, or $t-3$, 
versus use the SLOW predictor if the policy changed.
(See its CPT in Figure~\ref{fig:CPT}). 
Figure~\ref{fig:SIMLR_Results}(a) shows a change in the trend of reported new cases at week 22. 
However, just by looking at the evolution of number of new infections before week 22, there is no way to predict this change, 
which is why tf-v-SIR predicts that the number of new infections will continue growing. 
However, since \SIMLR\ observed a change in the government policies at week 20,
it realized it could no longer rely on its estimation of parameters and so switched to the SLOW model,
which is why it was more accurate here.
A similar behaviour occurs in week 34, when the 3rd wave of cases in Alberta started. 
Due to a relaxation in the policies on week 31, 
\SIMLR\ (at week 31) correctly predicted a change of trend around weeks 33--35.

Predictions at the country level are more complicated, 
since most of the time policies are implemented at the state (or province) level instead of nationally. 
For making predictions for an entire country, as well as predictions 3 or 4 weeks in advance,
\SIMLR\ first predicts, then uses, 
the likelihood of observing a change in trend, at every week. 
In these cases, the random variable $CT_{t+1}$ no longer acts like a ``switch'',
but instead it mixes the predictions of the tf-v-SIR and SLOW models, according to the probability of observing a change in the trend.

Figure~\ref{fig:SIMLR_Results}(b) shows that whenever there is a stable trend 
in the number of new reported infections
-- which suggests there has been no recent policy changes --
\SIMLR\ relies on the predictions of the tf-v-SIR model;
however, as the number (and rate of change) of new infections increases, so does the probability of observing a change in the policy.
Therefore, 
\SIMLR\ starts giving more weight to the predictions of the SLOW model. Note this behaviour in the same figure during weeks 13 -- 20.

Figure~\ref{fig:SIMLR_Results}(c) shows how our proposed \SIMLR\ approach compares with the 18 models that submitted predictions to the CDC during the same span of time. 
Note that \SIMLR\ and the model \emph{LNQ-ens1} are the best performing models, 
with no statistically significant difference (p > 0.05 on a paired t-test) with respect to MAPE. 
A striking result is how hard it is to beat the simple SLOW model 
(presented as \emph{COVIDhub-baseline} in the CDC dataset). 
Out of the 19 models considered here, only 5 (including \SIMLR) do better than this simple baseline when predicting 3 to 4 weeks ahead! 
This brings some insight into the challenge of making accurate prediction in the medium term -- probably due to the need to predict, then use, policy change information.

\section{Conclusions} 
\label{sec:concl}

\subsection{Limitations and societal impact} 
\label{sec:limitations}
We do not foresee any potential negative societal impacts of this work. There are, however, two main limitations to \SIMLR.
\SIMLR's underlying PGM makes it easy to interpret.
However, this structure does not imply causality:
Although changes in the observed policy influence changes in the trend of new reported cases, the opposite is also true in reality. 
Also, while \SIMLR\ is able to estimate the likelihood of a change in policy, it does not identify which policy might be implemented next.

The second limitation is that \SIMLR\ relies on 
conditional probabilities tables that are hard to learn
due to lack of data,
which forced us to build them based on domain knowledge. 
If these estimations are disproportionately wrong, then the predictions might be also misleading. Also, different regions might have different ``thresholds'' for taking action. Although for our experiments we set those CPTs a priori for US and Canada, they might differ for other countries.

Despite these two limitations,
\SIMLR\ produced state-of-the-art results in both forecasting in the US as a country, and  at the provincial level in Canada. 

\subsection{Contributions} 
This paper demonstrated that a model that explicitly
models and incorporates government policy decisions
can accurately produce 1- to 4-week forecasts of the number of  COVID-19 infections.
This involved showing (1)~that an SIR model with time-varying parameters is enough to describe the complex dynamics of this pandemic, 
including the different waves of infections. 
Modelling \SIMLR\ as a probabilistic graphical model not only makes it interpretable, 
but also makes it easy to incorporate domain knowledge that compensates for the relatively scarce data. 
\SIMLR's excellent performance 
-- comparable to state-of-the-art systems in this competitive task -- 
show that it is possible to design interpretable machine learning models without sacrificing performance.

We expect that this approach will be useful not only for modelling COVID-19, but other infectious diseases as well. 
We also hope that 
its interpretability will leads to its adoption by 
researchers, and users, in epidemiology and other non-ML fields.

\bibliography{main}
\bibliographystyle{abbrvnat}

\end{document}